\pdfoutput=1

\documentclass[11pt]{article}

\usepackage[]{acl}

\usepackage{times}
\usepackage{latexsym}
\usepackage{graphicx}
\usepackage{amsmath}
\usepackage{bbm}
\usepackage{listings}
\usepackage{booktabs}
\usepackage{comment}
\usepackage{xspace}
\usepackage[disable]{todonotes}
\usepackage{dialogue}
\usepackage{arydshln}

\definecolor{dkgreen}{rgb}{0,0.6,0}
\definecolor{gray}{rgb}{0.5,0.5,0.5}
\definecolor{mauve}{rgb}{0.58,0,0.82}

\definecolor{agent}{RGB}{119, 169, 204}
\definecolor{human}{RGB}{221, 180, 188}
\definecolor{llm}{RGB}{165, 155, 174}

\usepackage[T1]{fontenc}

\usepackage[utf8]{inputenc}

\usepackage{microtype}

\newcommand{\iws}{ChatShop\xspace}

\title{\iws: Interactive Information Seeking with Language Agents}

\author{Sanxing Chen\quad Sam Wiseman \quad Bhuwan Dhingra \\
  Duke University \\
  \texttt{sanxing.chen@duke.edu} \\
  \texttt{\{swiseman, bdhingra\}@cs.duke.edu} \\}

\begin{document}
\maketitle
\begin{abstract}
  The desire and ability to seek new information strategically are fundamental to human learning but often overlooked in current language agent evaluation. We analyze a popular web shopping task designed to test language agents' ability to perform strategic exploration and discover that it can be reformulated and solved as a single-turn retrieval task without the need for interactive information seeking. This finding encourages us to rethink realistic constraints on information access that would necessitate strategic information seeking. We then redesign the task to introduce a notion of task ambiguity and the role of a shopper, serving as a dynamic party with whom the agent strategically interacts in an open-ended conversation to make informed decisions. Our experiments demonstrate that the proposed task can effectively evaluate the agent's ability to explore and gradually accumulate information through multi-turn interactions. Additionally, we show that large language model-simulated shoppers serve as a good proxy for real human shoppers, revealing similar error patterns in agents.

\end{abstract}

\section{Introduction}
Humans consistently acquire new information and maintain their knowledge to make informed decisions~\cite{schmidhuber2010formal}. In interactive settings, these two aspects often intertwine to form a sequential decision-making process, in which the agent must decide how to act based on current knowledge or how to gather more information through efficient exploration. In order to achieve human-level learning, it is crucial for artificial intelligence language agents to possess similar abilities.

\begin{figure}[t]
  \centering
  \includegraphics[width=\columnwidth]{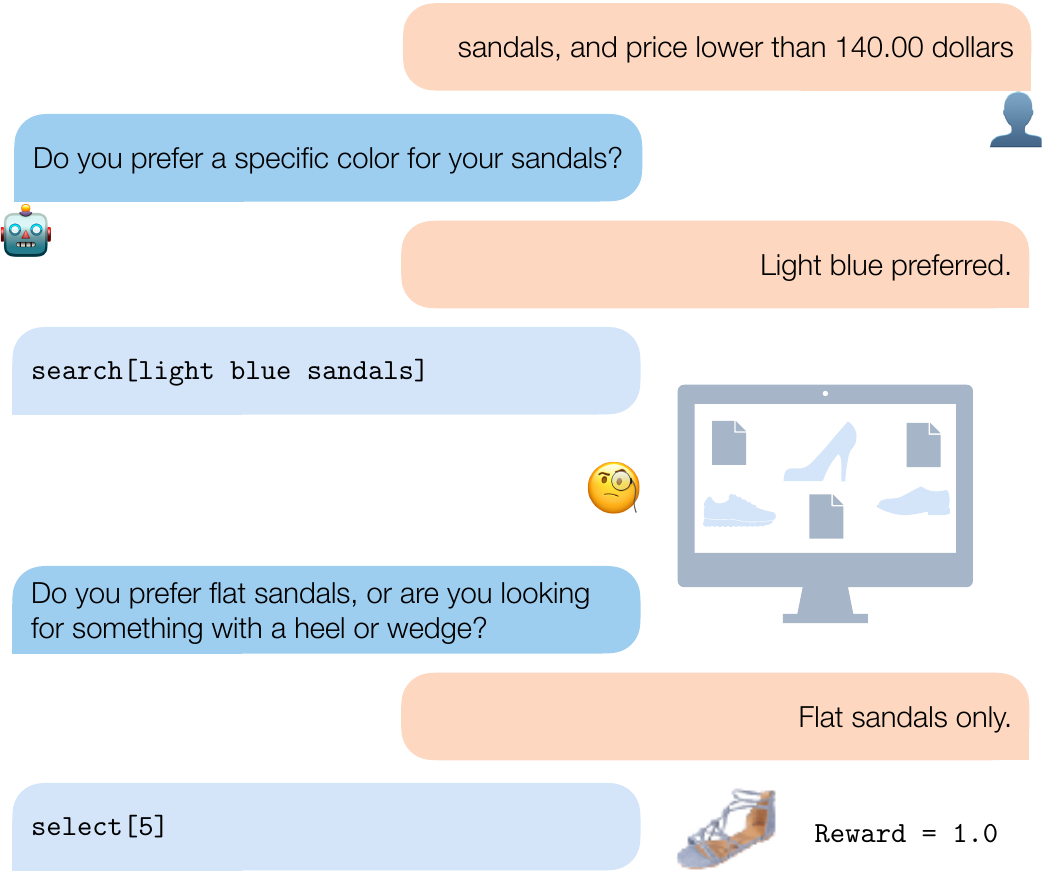}
  \caption{An illustration of the \iws task. In contrast to WebShop's original detailed instruction \textit{a pair of non-slip sandals for my wife that is blue in color, size 5.5}, we provide only the product type and price. We require the agent to narrow down the search space by interacting with the shopper and product database.}
  \label{fig:iWebShop}
  \end{figure}

  Recent studies have explored Large Language Models (LLMs) as autonomous agents in general problem solving~\cite{zhou2023webarena,liu2023agentbench,xie2023openagents}. In their design, the information-seeking component often applies to static information sources such as knowledge graphs or a collection of web documents in databases. The unrestricted access to these information sources reduces task \textit{interactivity}. The agent can skim through the entire information space without needing an efficient exploration strategy. 

  As a result, it remains unclear how strategic language agents are at seeking new information and whether their decision-making process is informed by tracking accumulated information. To investigate this issue, we first examine a popular web shopping task~\cite{yao2022webshop}. This task is designed to evaluate the agent's ability to strategically navigate a web interface and identify the correct product given a detailed goal instruction. 
  
  While WebShop constrains access to static information sources by layering UI interfaces, we find that the homogeneous interaction nature of this artificial constraint allows it to be automated by simple heuristics. Thus, with both the full information of the candidates and the full details of the target product available in the shopper's instructions, a retrieval system can directly score the relevance of each candidate to locate the target in one step without the need for strategic information seeking.

  In a real-world shopping scenario, a shopper likely does not possess full information about the target product and starts browsing alternatives with partial information~\cite{chen2009information,darley2010toward}. Their decision becomes clearer during this process. An automatic agent can help them better understand the product space and figure out their preferences through conversation. Introducing a shopper role in this task not only makes it more natural but also creates a natural constraint on information access. As a dynamic party, the shopper can exhibit a variety of behaviors, and a good user experience requires the agent to ask questions strategically and efficiently. 

  The key challenge in designing such a setup is that interaction between the agent and the shopper requires a human-in-the-loop environment, hindering scalable evaluation. Given the strong performance of recent LLM agents, we hypothesize that LLMs themselves are capable of simulating a human-like party in an interactive web shopping experience~\cite{li2023metaagents,park2023generative}. To test this hypothesis, we repurpose WebShop to propose \iws, in which the agent starts with an unspecific goal instruction—only the coarse type of product.
  This lack of specificity in the instruction creates a challenge of task ambiguity~\cite{tamkinHSG23}, which can only be resolved by effectively gathering information from the shopper and the website environment about products (\autoref{fig:iWebShop}). The challenge is amplified by other inherent complexities such as searching the vast product space and tool usage.

We benchmark a range of agents using both GPT-3.5/4 and a Llama 2 variant as base models in environments where either humans or LLMs play the role of the shopper. Experimental results verify the challenges introduced by the information need and constrained communication channels. We explore standard prompting strategies and simple heuristics to incentivize the agent. Our findings show that open-source models still lag behind commercial ones and, perhaps surprisingly, that a stronger base model does not always guarantee better performance in information-seeking tasks. We further evaluate how well an LLM simulates interaction with real human shoppers in a human study. The benchmarking results and failure patterns indicate that the LLM-simulated environment is effective in recovering the gap between agents. We hope our work can drive the automatic evaluation of language agents towards more complex and meaningful interactions with (simulated) humans.\footnote{Data and code will be shared upon publication.}

\section{Related Work}

\subsection{Information Seeking Tasks}

Language agents' information-seeking ability has long been a focus of AI research, especially in the context of question answering and task-oriented dialogue~\citep{dhingra-etal-2017-towards,zamani2022conversational,zhou2023webarena}.
In such tasks, the agent usually receives an information need from the user and accesses external knowledge sources to gather information,
a task which
can often be formulated as a single-turn retrieval problem.
The constraints imposed on such interaction are often artificial~\citep{yuan-etal-2020-interactive-machine}.
In contrast, the constraints in \iws task originate from a realistic situation of interacting with a human party in a web shopping scenario.

Reference games such as the popular twenty questions game\footnote{\url{https://en.wikipedia.org/wiki/Twenty_questions}} have long been studied as a testbed for information-seeking dialogue~\citep{bachman2016towards,guesswhat_game,guesswhich_game}.
Most reference games are either limited to yes-no questions over a
narrow, well-defined set of attributes.
\iws simulates a more realistic scenario where the agents are tested on the ability of asking open-ended questions and utilizing non-binary feedback.\footnote{In fact, finding the smallest set of attributes that can distinguish a target object from distractors can be viewed as a special case of the \textit{set cover problem}, which is already known to be NP-hard~\citep{dale1995computational}.}
In the context of reference games, past work has studied the generation of informative clarification questions~\citep{qi-etal-2020-stay,white-etal-2021-open} and the role of listener modeling~\citep{singh-etal-2023-know}.

\subsection{Human-AI Collaboration}

More recently, there has been a growing interest in studying human-AI collaboration via LLMs.
MINT~\citep{wang2023mint} benchmarks a range of LLM agents in leveraging human or AI-simulated feedback to improve multi-turn problem solving.
Unlike \iws, this feedback can be viewed as a form of natural language supervision, which is beneficial but not required to solve the task.
DialOp~\citep{lin2023decisionoriented} focuses on the agent's ability of planning based on human preferences in a grounded dialogue setting.
Compared to \iws, the tasks in DialOp has a narrower and synthetic search space.
\citet{li2023eliciting} propose a learning framework for LLMs to elicit human preferences in tasks such as content recommendation,
effectively leveraging LLMs' information seeking capabilities to reduce task ambiguity.
However, their tasks focus on exploration guided by the general world knowledge stored in the LLM weights internally, whereas in \iws, the exploration is grounded in an external real-world product space.

\section{Retrieval Approach for Webshop}
\label{sec:webshop}
In this section, we review the background of the original WebShop task~\cite{yao2022webshop}. We then demonstrate that its challenge to strategic exploration is artificial and that a small ranking model largely solves this task.

\subsection{Background}
WebShop presents a web shopping scenario in which an agent is given a goal instruction (e.g., \emph{I want a noise-canceling Cosycost USB microphone}) and navigates a web interface to identify the correct product from more than a million candidates scraped from Amazon. The typical actions available in WebShop involve querying a sparse BM25 search engine, clicking on product details, and confirming a product with corresponding options specified in the goal instruction. The task emphasizes the challenge of recognizing product types, extracting common bi-gram attributes (e.g., \emph{heavy duty, dust proof}) from lengthy product descriptions, and matching options and prices from a vast collection of products. 

WebShop designs a reward function based on the title string similarity and attribute coverage of the selected product compared to the goal product used in the instruction annotation process. The performance of the agent is evaluated based on the reward of the final product selected and the success rate of finding the correct product (i.e., reward equals 1).

The product space in WebShop is organized into layers of information (i.e., search results page, item page, and item-details page) to create an interactive interface for the agent to explore. However, due to the homogeneous structure of the interface—where every product is first shown as a short title in a list and then detailed on a separate page—we can automate the information-seeking interaction by simply clicking on each product and recording the information. As a result, the task can be reformulated as a retrieval problem against a list of product descriptions.

Additionally, because the instruction is the only specification of the WebShop task, it is meant to be sufficiently informative for an agent to identify the correct product. Therefore, we hypothesize that \emph{the relevance of each product can be independently determined by the goal instruction alone}. 

\subsection{The Retrieval Solution}
As initial evidence of the above hypothesis, we find that using the instruction as the search query with the built-in BM25 search engine returns a list of 50 products containing a full reward product 86.8\% of the time. This finding largely voids the need for the agent to learn how to strategically engage the search engine as a tool and diminishes the challenge of large product space exploration.

We further validate the hypothesis by training a simple BERT-based ranking model on the list of 50 products retrieved using the goal instruction. This model applies a cross-attention mechanism between the goal instruction and the concatenated textual product information. It uses a pairwise margin loss to effectively distinguish suitable products from unsuitable ones.

\begin{equation}
  \resizebox{.9\hsize}{!}{$
L = \sum_{i=1}^{N}\sum_{j=1}^{N} \max(0, 1 + (1 - 2\mathbbm{1}\{y_i > y_j\}) (\hat{y}_i - \hat{y}_j))
  $}
\end{equation}

where $\hat{y}$ is the predicted similarity score between the goal instruction and a product, $y$ is the reward of the product, and $\mathbbm{1}$ is the indicator function. The similarity score between the instruction and the product is computed from the last layer \texttt{[CLS]} token representation of the BERT model, followed by a linear layer. In cases where the ground truth product is present in the retrieval list, we add an auxiliary cross-entropy loss to encourage the model to rank the ground truth product higher than the other products.

Using the retrieval approach, we achieve a 78.3\% success rate and an 87.2 average reward on the dev set. This result is superior to the reported 59.6\% success rate and 82.1 average reward of human expert annotators~\cite{yao2022webshop}.\footnote{We do not consider the option selection component of WebShop in this retrieval study, as it merely requires exact string matching.}
Our findings suggest that the challenge of WebShop primarily involves semantic matching between the instruction and the product description. The interaction with the website does not present significant challenges or require strategic planning.
This observation motivates us to rethink a task design that captures meaningful interaction in real-world shopping scenarios.

\begin{table}[htbp]
  \centering
  \begin{tabular}{lrr}
    \toprule
    & \textbf{WebShop} & \textbf{\iws} \\
    \midrule
    \# Vocab & 2871 & 1166 \\
    Avg. Length & 15.1 & 2.3 \\
    \bottomrule
  \end{tabular}
  \caption{Corpus statistics of the original and simplified goal instructions. We tokenize the sentences using the nltk library and ignore the stopword tokens in vocabulary counting.}
  \label{tab:instruction_stats}
\end{table}

\section{\iws}

WebShop adopts a setup where the shopper provides a goal instruction containing full details at the beginning of the process. This design not only obscures the need for strategic information seeking but also diverges from real-world shopping scenarios. It is widely acknowledged that consumers undergo three stages—problem recognition, information search, and evaluation of alternatives—before making a purchase decision~\cite{darley2010toward}. In \iws, we simulate such a process by starting with an underspecified goal and having the agent assist the shopper in becoming more informed about their needs. The agent communicates with the shopper by asking questions and presenting alternative products, and the shopper provides feedback whenever new information comes to light. This section describes how we repurpose WebShop to implement this design and evaluate the protocol of our \iws task.

\paragraph{Simplified Goal Instruction}
In \iws, we aim to create a starting point with limited information for the agent to explore and accumulate information. We achieve this by simplifying the goal instructions of WebShop to a basic description of the type of item. We hide all attributes and options of the target product, leaving their discovery to the agent. We process the 1500 goal instructions in the development and test sets of WebShop and obtain simplified instructions using GPT-3.5 with few-shot prompts. The simplified instructions are six times shorter and contain fewer unique tokens than the original instructions. This outcome suggests a greater degree of task ambiguity~(\autoref{tab:instruction_stats}).

\paragraph{Agent and Shopper}
The proposed \iws task involves two roles: a shopper with the intent to purchase an item and an agent that assists the shopper in finding the correct product. The shopper is a dynamic party in the environment who has access to the target item information and uses it to answer clarification questions from the agent. In contrast, the agent is responsible for the challenging work of exploring the product space and summarizing important product features that the shopper may care about in their purchase decision.

This setup poses requirements on the information-seeking capabilities of the agent in two aspects: 1) finding features that have the most discriminative power to distinguish the target product from distractors, which requires a structured understanding of the product space; and 2) learning shopper behavior regarding which types of questions are most likely to elicit informative answers. For instance, the shopper is more likely to have preferences over certain features than others.\footnote{We provide the shopper with a list of important attributes from WebShop, many of which are bi-grams.} Another example is that some open-ended questions might prompt the shopper to reveal critical information about the target product, while others might be too vague and bothersome to answer.

Ideally, the shopper role should be played by a real customer with an actual goal of purchasing the target product. However, this is impractical for development in an academic setting without a large user base. Thus, we simulate the shopper using a language model-based chatbot. Note that the goal of the simulated shopper is not to mimic human behavior perfectly, but to provide a realistically consistent interaction partner that can be strategically engaged. Meanwhile, a variety of language agents can be developed and benchmarked in the agent role.

\paragraph{Action Space}
Three actions are available to the agent:  1) \texttt{search[query]}. This action initiates a search using a BM25 search engine, which returns a ranked list of products. The sparse retriever indexes a concatenation of product information, including the title, description, and options.  2) \texttt{question[content]}. When more information is needed for a precise decision, the agent interacts with the shopper for further clarification.  3) \texttt{select[index]}. When a single product is determined, the agent finalizes its recommendation.

\paragraph{Communication Channel}
In this task, we investigate two types of interaction:  1) open-ended text-based interaction. The agent is allowed to ask open-ended questions, and the shopper responds naturally in text.  2) instance-based comparison. The agent presents an item to the shopper, and the shopper provides comments on the item by comparing it to the requirements of the target product. Since the shopper has knowledge of the exact target product, there is a risk of the shopper directly revealing the target product through any communication channels. To prevent this, we instruct the shopper to reserve information until prompted and limit the length of the shopper's response. Empirically, we find these restrictions to be effective.

\paragraph{Limit and Reward}
Since the cost and speed of modern search engines are highly optimized, we do not constrain or penalize the use of the BM25 retriever in \iws. However, interacting with the shopper can be costly because it requires human involvement. Inefficient or redundant questions can harm the shopping experience. Therefore, we limit the maximum number of rounds the agent can interact with the shopper in each session. At the end of each session, when a single product is selected, we apply the same reward function against the ground truth product as in WebShop to obtain a score for the session.

\section{Experiments}

In this section, we introduce the experimental setup for evaluating agents in \iws. We first describe the environment setup and the benchmarking agents. We then present the results of the agents under different settings of information disclosure and interaction strategies. Finally, we evaluate the effectiveness of the LLM-simulated shopper in a human study. Additionally, we include a study in \autoref{sec:rl} to characterize the challenges of applying traditional reinforcement learning methods to the \iws task.

\subsection{Environment Setup}

We use OpenAI's GPT-3.5 model to simulate the shopper for automatic evaluation. The simulated shopper is provided with the title, the required attributes, and the options of the target product in the prompt. The shopper is instructed to respond to the agent's questions using fewer than 5 words and under a token limit of 10. We allocate a question budget of 5 for each session. Unless specified otherwise, we assess the agent over 100 sessions. In practice, we observe that the agent's performance remains consistent with this number of samples.

\subsection{Benchmarking Agents in \iws}
We select three representative LLMs (OpenAI's GPT-3.5/4 and \textsc{CodeLlama}-32b) as the backbone of agents to be tested in our study.\footnote{We use gpt-3.5-turbo-1106 and gpt-4-1106-preview versions.} The complexity of this multi-turn task and the constrained context length of the LLMs make it impractical to include few-shot demonstrations in prompts. Consequently, we carefully design zero-shot prompts to instruct the agent to reason and generate valid actions situationally. Since the number of calls to the search engine is unlimited, repeated search actions can lead to a lengthy context for the agent to process and risk exceeding the context window limit. Therefore, we compress the conversation history by hiding the product candidate results from previous search actions so that the agent can focus on the current state of the game. The old search results are removed from the context and hidden from the agent when a new search is initiated. All prompts are listed in the supplementary material.

We implement three prompting strategies with heuristic action enforcement:
1) \textit{auto q}: the agent decides on its own whether to ask questions or search until it chooses to finalize the task with a product decision;
2) \textit{all q}: the agent performs a search engine query at the beginning and asks all possible questions until the budget is used up, then finalizes the task with a product decision;
3) \textit{interleave}: the agent asks questions and queries the BM25 search engine in an interleaved manner until using up all the questioning budget.
After that, the agent independently decides whether to initiate additional search queries or finalize the decision. This strategy is designed for a balanced utilization of the tool usage and questions.

To enforce these strategies, we use the \texttt{tool\_choice} configuration in the OpenAI API to control the agent's action selection. For open-source models, we use explicit cues in the prompt and resample the model's response until it follows the strategy.

\begin{table}[tbp]
  \centering
  \small
  \begin{tabular}{lrrr}
    \toprule
    Channel & CodeLlama & GPT-3.5 & GPT-4 \\
    \midrule
    None / Basic & 34.3 & 43.4 & 48.8 \\
    Open-ended & - & 40.6 & 49.7 \\
    Instance & - & 40.4 & 51.3 \\
    None / Full Info & 64.5 & 76.0 & 80.1 \\
    \bottomrule
  \end{tabular}
  \caption{Avg. rewards of (\textit{auto q}) agents under different settings of information disclosure. The two `None' settings are without interactions with the shopper. \textsc{CodeLlama} cannot perform under the interactive settings without advanced prompting strategies.}
  \label{tab:models_results}
\end{table}

\paragraph{Challenge of Information Scarcity}
In the results of \autoref{tab:models_results}, we first find that state-of-the-art LLMs generally achieve a high reward with access to full information in instructions (`None / Full Info'), which mimics the setting of WebShop. This aligns with our earlier observation that, with a flattened UI interface, the WebShop task is not particularly challenging for the agent to navigate. When we withhold the full information in instructions (`None / Basic'), all of the tested LLM agents perform significantly worse, with a performance drop of more than 30\% in average rewards. This confirms the information scarcity created by the simplified instructions in \iws, which necessitates that the agent actively seeks information from the environment.

However, when given access to interact with the shopper, the agents still struggle to utilize the communication channel effectively, resulting in a performance similar to or even lower than the no-interaction setting. We find that the \textit{auto q} prompting strategy is inadequate to incentivize the agents to interact with the shopper. The agents are often overconfident in making decisions based on partial information from the basic instruction or brief interaction with the shopper, despite being prompted to ask questions until ``\textit{the user's criteria clearly match a single product}''.

\begin{table}[tbp]
  \centering
  \small
  \begin{tabular}{lrrrrrr}
    \toprule
    Strategy & \multicolumn{2}{c}{\textbf{CodeLlama}} & \multicolumn{2}{c}{\textbf{GPT-3.5}} & \multicolumn{2}{c}{\textbf{GPT-4}} \\
    ReAct & w/o & w/ & w/o & w/ & w/o & w/ \\
    \midrule
    \textit{no q} & 34.3 & 30.1 & 43.4 & 45.6 & 48.8 & 47.5 \\
    \textit{auto q} & - & - & 40.6 & 62.7 & 49.7 & 59.2\\
    \textit{all q} & 25.6 & 29.4 & 63.7 & 61.3 & 63.0	& 66.3 \\
    \textit{interleave} & 18.8 & 28.9 & 64.3 & \textbf{68.2} & 60.5	& 68.1 \\
    \bottomrule
  \end{tabular}
  \caption{Avg. rewards of agents with different strategies and the open-ended communication channel. \textit{no q} is the non-interactive baseline.}
  \label{tab:strategy_results}
\end{table}

\paragraph{Advanced Prompting Strategy}
LLM agents have demonstrated limited ability to leverage the communication channel in the \textit{auto q} setting. We aim to determine whether stronger agents can be achieved through task heuristics and common prompt engineering techniques such as ReAct~\cite{react}. 
For ReAct, the agent is instructed to interleave its actions with a summary of the information gathered up to the current turn and reasoning for the next action. In \autoref{tab:strategy_results}, we observe that ReAct is significantly more effective in interactive settings, especially in the \textit{auto q} setting where the agent is otherwise confident and reluctant to ask questions.

In the optimal scenario, GPT-3.5 surprisingly outperforms GPT-4. This suggests that stronger base model performance does not always translate to superior performance in information-seeking tasks. Although advanced prompting strategies further incentivize the agents, a substantial gap remains between the best agent and the non-interactive full information baseline.

\begin{table}[tbp]
  \centering
  \small
  \begin{tabular}{lrrrr}
    \toprule
    Strategy & \multicolumn{2}{c}{\textbf{Open-ended}} & \multicolumn{2}{c}{\textbf{Instance}} \\
    ReAct & w/o & w/ & w/o & w/ \\
    \midrule
    \textit{no q} & 43.4 & 45.6 & 43.4 & 45.6 \\
    \textit{auto q} & 40.6 & 62.7 & 40.4 & 51.6\\
    \textit{all q} & 63.7 & 61.3 & 48.3	& 47.1 \\
    \textit{interleave} & 64.3 & \textbf{68.2} & 51.1	& 51.3 \\
    \bottomrule
  \end{tabular}
  \caption{Avg. rewards of the GPT-3.5 based agents with different interaction strategies and both open-ended communication channels. \textit{no q} is the non-interactive baseline.}
  \label{tab:strategy_results_instance}
\end{table}

We further investigate the effectiveness of the advanced prompting strategies in the instance-based communication channels in \autoref{tab:strategy_results_instance}. The results show that these strategies are generally helpful in improving the agent's performance, but the agents still struggle to utilize the communication channel effectively.

\subsection{LLM versus Human Shopper}
To understand the effectiveness of using LLMs as a reasonable proxy for simulating human-like shopper interactions, we compare the performance of LLM agents with simulated shoppers to that of real humans. 
We recruit 8 participants to play the role of the shopper in the human study. Each participant is asked to complete 10-20 sessions of the \iws task. The average completion time for one session is 2.5 minutes.\footnote{OpenAI API wait time accounts for about 30\% of the total time.}
We compare two OpenAI agents in the study, both utilizing the \textit{interleave} strategy.
Additionally, we allow the GPT-3.5 agent to employ ReAct style reasoning.\footnote{GPT-4 is slow to generate responses with ReAct style reasoning, which is not suitable for the human study.}
The prompts are revised for a better human experience, providing instructions about not asking for pricing information and avoiding vague questions. This revision results in lower performance of GPT-3.5 compared to the original setting.
In total, we collect data from 100 human shopping sessions.

\begin{table}[tbp]
  \centering
  \begin{tabular}{lrr}
    \toprule
    & \textbf{GPT-3.5} & \textbf{GPT-4} \\
    \midrule
    Simulated & 59.0 & 62.8 \\
    Human & 58.2 & 63.4 \\
    \bottomrule
  \end{tabular}
  \caption{Avg. rewards of LLM agents with simulated and human shoppers over 50 sessions.}
  \label{tab:human_results}
\end{table}

From the results in \autoref{tab:human_results}, we find that the LLM agents' performance with both the simulated shopper and the human shopper is consistent. Both environments present a similar level of challenges to the agents and reveal the gap between the two agents.

\begin{table}[tbp]
  \centering
  \begin{tabular}{lcc}
  \toprule
   & Human & LLM \\
  \midrule
  Search query length & 71.1 & 73.7 \\
  Precision@5 & 19.8 & 18.8 \\
  Avg reward@5 & 51.5 & 53.6 \\
  Hit@5 & 42.0 & 43.0 \\
  \bottomrule
  \end{tabular}
  \caption{\label{tab:search_perf} BM25 Search Performance: Human vs. LLM-generated shopper responses}
  \end{table}

  We further look into the comparative performance of large language models (LLMs) and human shoppers in terms of response quality. Specifically, we are interested in quantifying the amount of task-specific information conveyed by each source when answering the same set of questions. This setup contrasts with the previous end-to-end analysis of end task performance where agents may choose different trajectories (i.e., a different set of question) based on responses.

  To facilitate this comparison, we first generate simulated shopper responses to the set of 500 questions (from the 100 session/trajectories) for which we already had human response data. Subsequently, we employ GPT-4 to summarize the list of question-answer (QA) pair belonged to each session into a concise search query. These queries are then utilized to retrieve results using the BM25 search engine.
  
  Our findings are summarized in the comparative metrics presented in~\autoref{tab:search_perf}.
  From these results, we infer that the LLM shopper tends to provide slightly more information per query, evidenced by the marginally longer search query length and higher average reward. However, the differences in precision and hit rate between human and LLM shoppers are minimal, indicating that while LLM shoppers may offer more extensive information, this does not substantially enhance end-task performance compared to human responses.

  \begin{table}[tbp]
    \centering
  \begin{tabular}{|p{0.95\linewidth}|}
    \hline
    Product Name: Gold Label Men's Big and Tall Non-Iron Point Collar Dress Shirt Wrinkle-Resistant Long Sleeve \newline
    Attributes: long sleeve, daily wear \newline
    Options: lt lavender 012 (pink), 20" neck 34" sleeve, big \newline
    \textit{The above information is hidden from the agent.}\\ \hdashline
    
    Goal: shirt \newline
    Budget: 80.0 \\ \hdashline
    
    \textcolor{agent}{Agent}: Do you prefer a shirt with long sleeves or short sleeves?
\newline
\textcolor{human}{Human}: long sleeves
\newline
\textcolor{llm}{LLM}: Long sleeves.
\newline
\textcolor{agent}{Agent}: Do you prefer a shirt for men or women?
\newline
\textcolor{human}{Human}: men
\newline
\textcolor{llm}{LLM}: Men's shirt.
\newline
\textcolor{agent}{Agent}: Do you have a preference for color in your long-sleeve men's shirt?
\newline
\textcolor{human}{Human}: pink
\newline
\textcolor{llm}{LLM}: Light lavender is preferred.
\newline
\textcolor{agent}{Agent}: Do you prefer a shirt with a plain design or with a pattern like cashew nut print or map print?
\newline
\textcolor{human}{Human}: plain design
\newline
\textcolor{llm}{LLM}: Plain design.
\newline
\textcolor{agent}{Agent}: Do you prefer a V-neck or crewneck style for your men's long sleeve pink shirt?
\newline
\textcolor{human}{Human}: i don't know
\newline
\textcolor{llm}{LLM}: Point collar only. \\
    \hline
    \end{tabular}
    \caption{\label{tab:dial} A dialogue example between a GPT-4 agent and both the human and LLM shoppers.}
    \end{table}

    Qualitatively, we observe that human shoppers sometimes fail to capture the relevant information from the product description, while LLMs remain more consistent and faithful. \autoref{tab:dial} demonstrates an example where the LLM shopper provides more informative responses than the human shopper. The more accurate description of color preference and the association of neck style with collar type align with our expected behavior. Overall, the LLM shopper is comparable to human shoppers in terms of the quality of information provided.

\subsection{Error Analysis}

To further understand the current limitations of LLM agents and the challenges of the \iws task, we conduct an error analysis on the failed trajectories. We categorize the failure patterns into five types, associated with systematic limitations of LLMs.

We define the five error types as follows:
\begin{enumerate}
  \item \textbf{Reversion}: The agent loses track of shopper-specified requirements. In the context of LLM agents, this is often caused by the agent's inability to robustly recall information across long contexts~\citep{liu2023lost}.
  \item \textbf{Misinterpretation}: The agent fails to understand the shopper's specified requirements. In our realistic shopping scenario, the task covers a diverse range of products and attributes, making grounded understanding of the shopper's intention challenging and error-prone.
  \item \textbf{Insufficient information gathering}: The agent does not gather enough information to locate the correct product, causing important attributes or options to be missing. This error is associated with the agent's lack of strategic information-seeking and overconfidence in making decisions based on partial information.
  \item \textbf{Repeated questions or search}: The agent asks the same question or searches the same query repeatedly, leading to inefficient actions. Language models are known to have a tendency to repeat themselves in long contexts~\citep{holtzman2020curious}.
  \item \textbf{Misleading user}: The shopper makes mistakes, being inconsistent or unclear. In a dynamic and interactive environment, it is natural for the shopper to make mistakes or correct themselves. The agent should be able to tolerate a certain level of noise and handle these cases gracefully. This also serves as a sanity check for the simulated shopper.
\end{enumerate}

Manually reviewing the lengthy trajectories of LLM agents is time-consuming and error-prone. Therefore, we adopt an automatic evaluation method by prompting GPT-4 to tag failed trajectories with the likely causes of failure as a multi-label classification problem. We manually verify a small subset of the model's predictions and find them consistent with our error type definitions.

\begin{figure}
  \centering
  \includegraphics[width=\columnwidth]{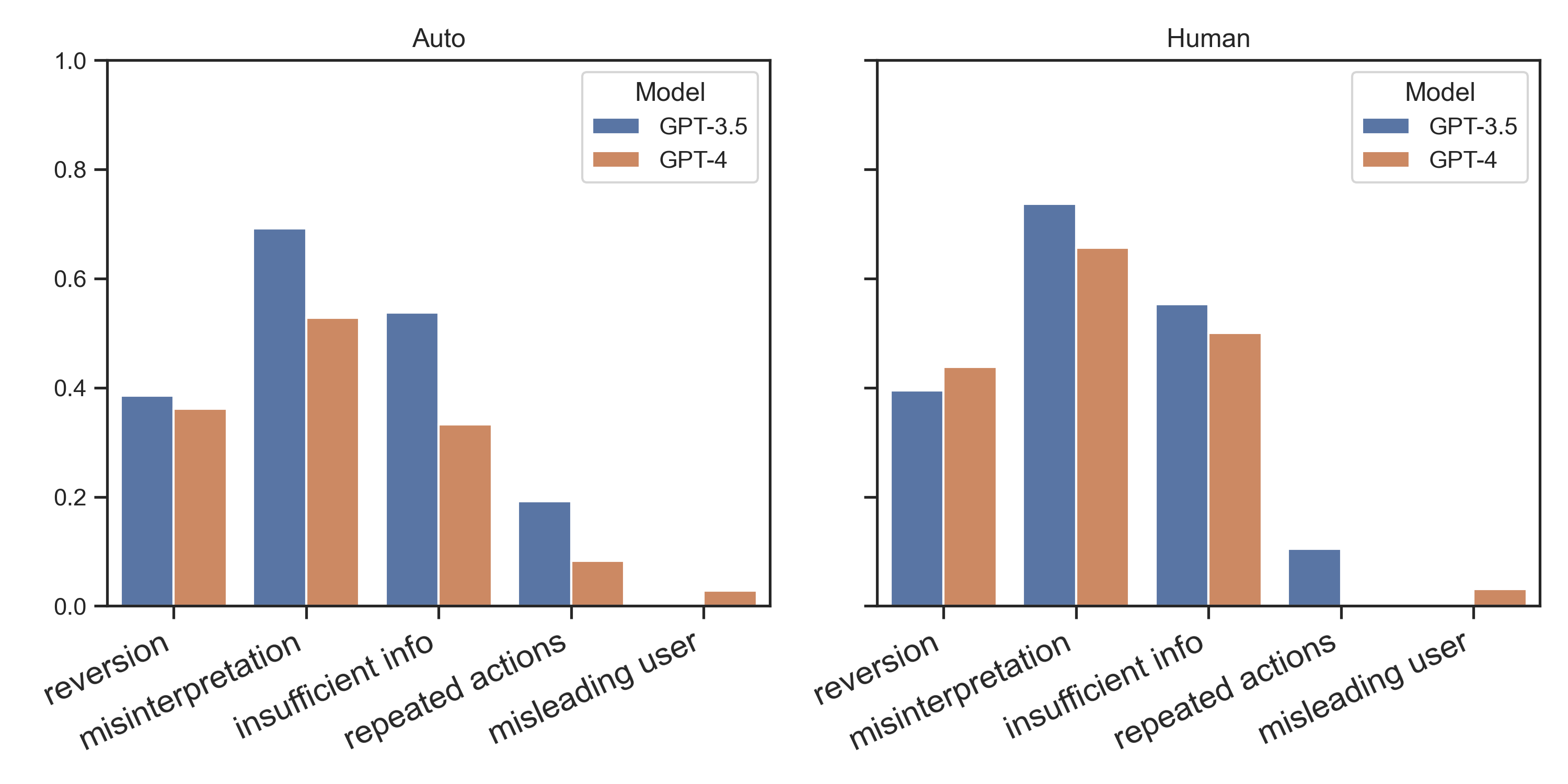}
  \caption{Relative frequency of error types in the LLM agents' failed trajectories with simulated and human shoppers.}
  \label{fig:error_dist}
\end{figure}

The results in \autoref{fig:error_dist} show that \textit{Misinterpretation}, along with \textit{Insufficient information gathering} and \textit{Reversion}, are the most common error types. This indicates the current LLM agents' lack of strategic information seeking and robust long context modeling. The inferior GPT-3.5 agent exhibits a higher rate of major error types. This gap is even more pronounced in the simulated shopper environment.

We compare the distribution of failure patterns between the two environments and find that the automatic environment successfully discovers a similar pattern of error types as observed in the human environment. The occurrence of the \textit{misleading user} error is rare in both environments, which suggests a robust environment and task design.

\section{Conclusion}
\iws presents a information-seeking centric evaluation of language agents, revealing a range of limitations of current LLM models.
We hope our fully automatic evaluation pipeline and baseline agents can benefit future research in creating agents that can explore and learn from the environment effectively.

\section{Limitations}
Our \iws task is realistic in the vast real product space and in its interaction with the shopper. However, it remains a simplified version of real-world web shopping scenarios. One unrealistic assumption is that the target product is known to the shopper during evaluation. Both human and LLM shoppers' behaviors can differ from real-world scenarios where the shopper is uncertain about the target product. Relaxing this assumption would require real-world data on the shopper's evolving knowledge of the target product during the search.

Under our current evaluation protocol, agents are evaluated based on end-task performance within a fixed budget of questions. This does not capture the quality of interaction for successful sessions, as they all receive full rewards. Future work can explore dynamic budget allocation strategies based on the difficulty of individual sessions or introduce penalties for asking uninformative questions.

Multimodal information, such as images and reviews, can also be integrated into the \iws task to provide a richer environment for the agent to explore.

\bibliography{anthology,custom}

\clearpage
\appendix

\section{Reinforcement Learning in \iws}
\label{sec:rl}

To characterize the challenge of applying traditional reinforcement learning (RL) methods to the \iws task, we conduct a preliminary study on a simplified version of the task named \textit{ChatShopBin}.

\subsection{ChatShopBin}
In \textit{ChatShopBin}, the agent receives no goal instruction but a list of products represented by their textual descriptions. The agent is required to interact with the shopper to identify the target product from the list. The action space of the agent is simplified to two actions: 1) \texttt{question[content]}: the agent asks about the existence of a certain attribute in the target product; the attribute is always a uni-gram such as `blue', `small'; 2) \texttt{select[index]}: the agent finalizes the task with a product recommendation. The shopper only provides binary feedback on the existence of the attribute in the target product.

One challenge we intend to study with ChatShop is for the agent to learn shopper behavior regarding what types of questions are most likely to be answered informatively through the interaction. In ChatShopBin, we define a simple, consistent shopper behavior: the shopper can only recognize attributes within a certain range of lengths. For attributes that are too long or too short, the shopper will always respond with a negative answer.

\subsection{RL Setup}
Following previous work on reinforcement learning agents for text-based games~\citep{he-etal-2016-deep,tuyls2022multi}, we train a Q-learning agent with a pre-trained BERT encoder as the backbone.
A game state of \textit{ChatShopBin} includes all product descriptions, a conversation history consists of the agent's questions and shopper's responses, and the current budget of questions.
The BERT encoder takes the concatenation of all components in the game state as input and generates the representation of the state.
We add a two-layer feedforward network with ReLU activation on top of the BERT encoder to predict a score for each token in the input.
As the action an agent can take is limited to words in the product descriptions, we take the scores corresponding to product description tokens as the Q-values of the agent's actions.
We append a special token to the beginning of each product description to represent the action of recommending the product.
Each question action comes with a cost of 1, 
the session ends when the agent finalizes the task with a product recommendation and the agent receives a reward of 5 for a successful session and -3 otherwise.

The training starts with a random policy and 8 parallel environments. We prioritize trajectories with the maximum reward in a replay buffer of size 100,000. The priority fraction is set to 0.5 meaning that half of the samples are still randomly sampled.
\begin{equation}
  \mathcal{L}_{\mathrm{TD}}(\phi)=\left(r+\gamma \max _{a^{\prime} \in A} Q_\phi\left(o^{\prime}, a^{\prime}\right)-Q_\phi(o, a)\right)^2
  \end{equation}
The agent is trained with the temporal difference loss and the discount factor $\gamma$ is set to 0.8. We use the Adam optimizer with a learning rate of $1e-5$ and a batch size of 64.

We also experiment with a policy gradient agent with the same architecture.
The agent is trained with the REINFORCE algorithm with a value baseline approximated by the average returns of recent episodes.

We are able to train agents that can achieve a near 100\% success rate and an average reward near 2.0 on the ChatShopBin task with both Q-learning and policy gradient methods, however, the training process is unstable (sensitive to hyper-parameters) and requires a large number of samples to converge.

\subsection{Potential-based Reward Shaping}

To further investigate the slow learning process of the RL agents, we compare our ChatShopBin task with conventional text-based games where similar but simpler agents have been successfully trained~\citep{tuyls2022multi}.
One of the key differences we identify is the lack of itermidiate reward signals in ChatShopBin.

The reinforcement learning setup we adopt utilize a prioritized experience replay buffer to focus on transitions that lead to the heighest rewards.
This is a common strategy for many text-based games where dependencies between actions are strong and random success is rare.
However, in ChatShopBin, the agent can randomly guess the target product with a success rate of 1/5 and receive the highest reward of 4.0.
This allows many unlearnable transitions (shortcuts) to be generated and prioritized during initial random exploration creating noises that greatly hinder the learning process.
Moreover, many text-based games have intermediate rewards that guide the agent towards the final goal, but in ChatShopBin, the agent can only receive positive rewards at the end of the session.
The total rewards from start is completely disassociated with the game progress (i.e., the agent is not rewarded when it secures important information).
Therefore, the agent must rely on intrinsic indicators of information to determine the learnable transitions and progress the game.

One solution to this issue is to reward the agent for information gain in the form of potential-based reward shaping~\citep{ng1999policy}.
Prior work has adopted potential-based reward shaping to accelerate learning for agents on a 20 Question Game conversational game~\cite{zhao-eskenazi-2016-towards}.
The information gain in their setup can be easily measured by the reduction in the number of candidates returned from a database.
However, in ChatShopBin, whether a candidate is ruled out or not is not directly observable, as the shopper behavior is unkonwn to the agent.
We therefore propose to learn an explicit listener model that captures the shopper's behavior and use it to estimate the validity of a candidate product based on the conversation history.

The listener model is based on a pre-trained BERT encoder taking as input the conversation history and the product description.
It simply predicts the probability of a candidate product being ruled out by the up-to-date feedback from the shopper.
The listener model is trained on trajectories that lead to successful sessions where we know for sure the target product is selected.
The target product and the conversation history are used as positive samples, and we sample negative samples by flipping some responses in the conversation history.
In a interleave manner, the listener model is trained with the agent.
We update the listener model every 1000 steps of the agent training and reset the environments to avoid mixed potential-based reward being added to one trajectory and lead to erroneous high rewards.
With this optimization, we are able to train both q-learning and policy gradient agents that converge fast and stable.

\subsection{Generalization to \iws}
Through the study of ChatShopBin, we identify the challenges of applying RL to the \iws task lie in sample efficiency and listener modeling.
Real-world shopper behavior is complex and dynamic, the feedback could be nonbinary and bear uncertainties.
We leave the exploration of more sophisticated and generalizable listener-aware RL agents for future work.

\begin{figure*}[t]
  \centering
  \includegraphics[width=6in]{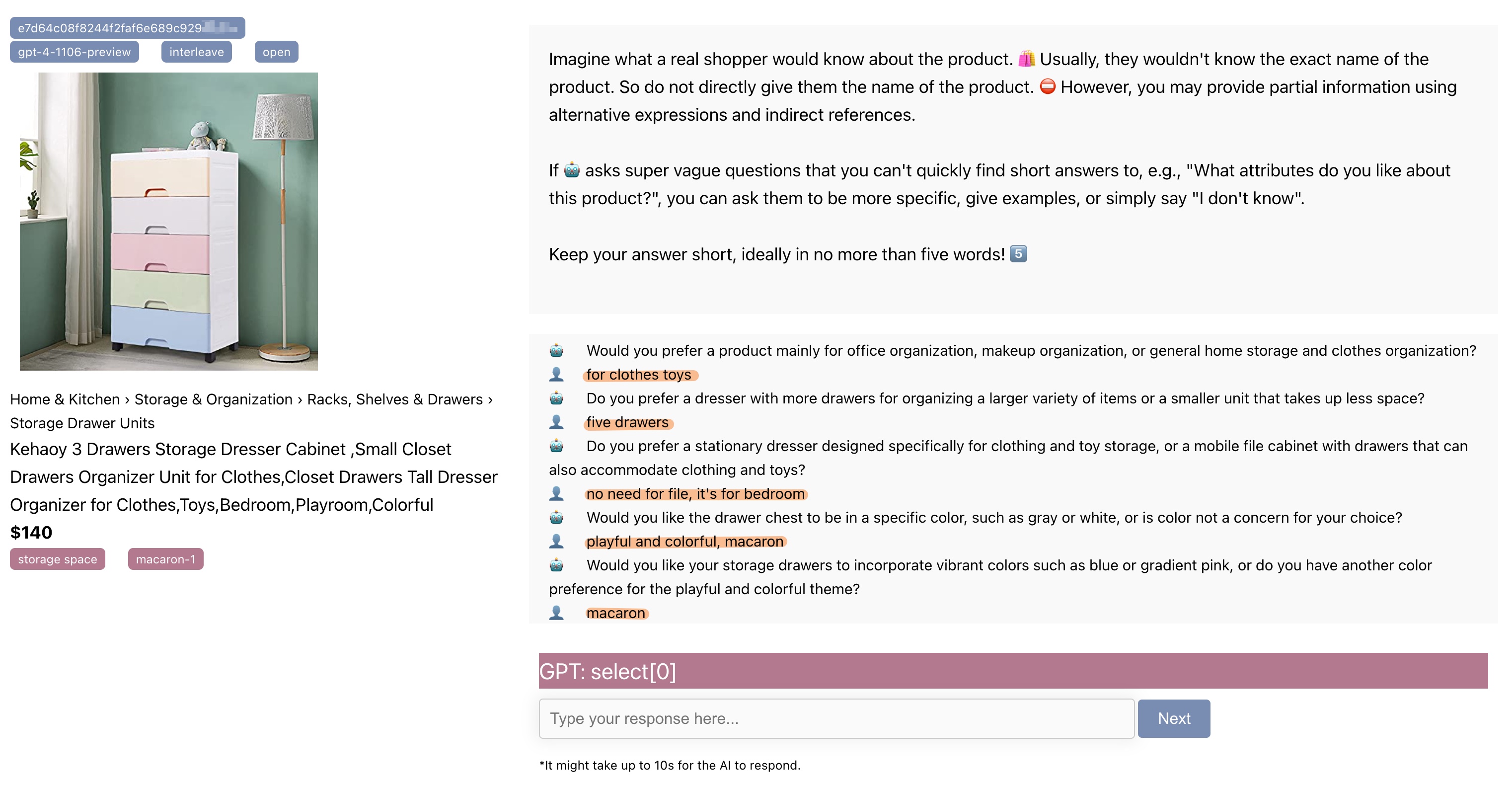}
  \caption{A GPT-4 agent helps a human shopper in the \iws task. Picture shows the web interface we build for human evaluation. The left panel provides shopper-related information such as the target product details. The right panel includes the goal instruction and a chat agent interface. The agent can ask questions to the shopper to gather information about the target product. The shopper is asked to answer within a certain length, thus limiting the information transmitted in a single interaction turn.}
  \label{fig:human_eval}
  \end{figure*}

\section{Experimental Details}

\subsection{Data Preparation}
\label{sec:data_prep}
We use the GPT-3.5 model to extract the coarse product type from the original WebShop goal instructions.\footnote{The WebShop dataset is MIT licensed and can be accessed at \url{https://github.com/princeton-nlp/WebShop}.}
The corpus statistics of the 1,500 (1,000 test, 500 dev) original and simplified goal instructions are shown in \autoref{tab:instruction_stats}.
We maintain the same training, development, and test splits as defined in the WebShop task.
As the agents presented in this study do not require training, we only evaluate and report their performance on the first 100 examples from dev set.

\subsection{Prompt Engineering for Agent}

We access the OpenAI models via paid APIs.
We host \textsc{CodeLlama} with Hugging Face's transformers library and query via the fastchat~\citep{zheng2023judging} API wrapper.
These three models are claimed to support long input contexts of at least 16k tokens.\footnote{We find that the 4,096 tokens context length of many LLMs (e.g., \texttt{gpt-3.5-turbo-0613} and \textsc{Llama} 2) is not sufficient for the task.}

In the zero-shot prompts, the agents see a brief description of the task and the goal instruction, and are instructed to ask questions to the shopper to gather information about the target product until the shopper's criteria clearly match a single product.
Through the conversation, the agents may choose to search the product space using the BM25 search engine. A list of products is returned and the cancatenation of the product titles, attributes, and available options is presented to the agent.
For OpenAI models, we provide 20 products for each search action, and for \textsc{CodeLlama}, we provide 5 products because slower response time and inferior performance in long context modeling.

For OpenAI models, we use \texttt{tools} to mitigate the challenges in enforcing valid agent actions and parsing model generation.
For \textsc{CodeLlama}, we use textual prompts only and interactively add lexical hints (e.g., ``What your next search | select would be'') to enforce valid actions.
We use regular expressions to parse the agent's response.
We set the maximum length of the agent's response to 100 tokens, and sample responses until a parseable response is generated.

\subsection{Prompt Engineering for Shopper}

We use OpenAI's GPT-3.5 to simulate the shopper in the \iws task.\footnote{We use the \texttt{gpt-3.5-turbo-0613} version.}
The shopper is given the product title, the required attributes and options of the target product.
To prevent the simulated shopper from directly revealing the target product, we adopt a simple yet effective strategy---limiting the length of the shopper's response.
We instruct the shopper to answer the agent's questions in less than 5 words and also cut off the shopper's response if it exceeds 10 tokens.
This fixed length constraint also encourages the agent to ask questions in a more strategic manner, optimizing the information density of each interaction turn.
We additionally instruct the shopper to avoid revealing the target product voluntarily, but to provide answers in a way that is consistent with the target product.
In their response, the shopper reminds the agent the left budget of questions and rejects the agent's request if the budget is exceeded.

\subsection{Sample Conversation}

Here is a sample conversation about ``bookcase'' between the GPT-4 agent and the simulated shopper in the \iws task.

\begin{dialogue}

  \speak{Agent} Do you have a preference for a specific type of cabinet such as wine, file, or general storage?
  
  \speak{Shopper} General storage cabinet.
  
  \speak{Agent} Do you prefer a specific color for your general storage cabinet?
  
  \speak{Shopper} White preferred.
  
  \speak{Agent} Do you prefer a cabinet that is freestanding or one that mounts on the wall?
  
  \speak{Shopper} Freestanding.
  
  \speak{Agent} Do you need the cabinet to have a specific height or size to fit in your space?
  
  \speak{Shopper} Standard size is fine.

  \speak{Agent} Would you like a cabinet with doors, drawers, or both for your storage needs?

  \speak{Shopper} Doors only.
  
  \end{dialogue}

Another example regarding ``hair chalk''.

\begin{dialogue}

  \speak{Agent} Do you have any preference for the number of colors in the hair chalk set? Or are there any specific colors you're looking for?
  
  \speak{Shopper} 6 colors.
  
  \speak{Agent} Do you prefer the hair chalk to be easily washable or are you looking for something that might last a bit longer before washing out?

  \speak{Shopper} Washable hair chalk.
  
  \speak{Agent} Are you looking for hair chalk that comes with additional hair accessories or are you only interested in the hair chalk itself?
  
  \speak{Shopper} Hair chalk only.
  
  \speak{Agent} Do you prefer hair chalk that is specifically marketed for kids and teens, or does it not matter to you as long as the product meets your requirements?
  
  \speak{Shopper} Doesn't matter.

  \speak{Agent} Is there a specific brand or any additional attribute you are looking for in the hair chalk set?

  \speak{Shopper} Non-toxic and easy clean.
  
  \end{dialogue}

  \subsection{Human Evaluation}
  We recruit 8 participants (university students) to play the role of the shopper in the human study. Each participant is asked to complete half to one hour of the \iws task. The participants are compensated on average \$12/h for their time.

\subsection{Error Types Classification}
\label{sec:err_types}
For classification of error types, we use the GPT-4 model to tag failed trajectories with the likely causes of failure as a multi-label classification problem.
We design a prompt consists of the flattened conversation history, the agent selected product, the goal product, and fine-grained rewards (i.e., title similarity, attribute/option coverage seperately).
The GPT-4 model judges the relevance of each error type based on the textual description of them and the episode context.\footnote{We use the \texttt{gpt-4-0125-preview} version.}

\section{\iws Prompts}
\label{sec:prompts}

\subsection{Shopper Prompt}
\textbf{System Prompt:}
\begin{lstlisting}
  You are playing the role of a shopper. While interacting, avoid explicitly stating the name of the product you intend to purchase. However, if prompted for specific related information, you may provide descriptions using alternative expressions and indirect references.

Product name: OWYN - 100%

Important attributes: gluten free
\end{lstlisting}
\textbf{Sample User Prompt:}
\begin{lstlisting}
  Do you have any allergies?
\end{lstlisting}

\subsection{Agent Prompt}
\textbf{System Prompt:}
\begin{lstlisting}

  Your role is to guide users through an online shopping experience, helping them find products that best fit their needs. When a user specifies certain attributes, you analyze these to sift through the available products, based on detailed product descriptions. There are three key actions:
  
  - `search[query]`: At the start, and whenever necessary, you can initiate a search using the website's BM25 search engine. Price can't be searched. This search yields a list of products, each with a unique description and index number. You may perform this action multiple times to refine the search based on evolving user requirements.

  - `select[item_index]`: When the user's criteria clearly match a single product, you finalize your response with `select[]`. Here, `item_index` refers to the unique number of the identified product.
  
  - `question[question_content]`: When multiple products fit the user's described attributes, or when more information is needed for a precise decision, you narrow down the choices with `candidates[0, 1, 2]` for example, listing the indexes of potential matches. Concurrently, you should pose questions to the user for further clarification.
  
\end{lstlisting}
\textbf{Sample User Prompt:}
\begin{lstlisting}
  Goal: i need to find a small end table that is easy to assemble; pick a blue-coated steel frame that won't rust\nThe next action is
\end{lstlisting}

\subsection{Subject Extraction Prompt}
\textbf{System Prompt:}
\begin{lstlisting}
  You assist users in extracting the main target from their search queries by removing all product attributes. Your response only contain the cleaned query.
\end{lstlisting}
\textbf{Sample User Prompt:}
\begin{lstlisting}
  User Query: "i want a noise cancelling cosycost usb microphone"
\end{lstlisting}
\textbf{Sample Assistant Prompt:}
\begin{lstlisting}
  microphone
\end{lstlisting}

\subsection{Attribute Removal Prompt}
\textbf{System Prompt:}
\begin{lstlisting}
  You assist users in refining their search queries by removing specific product attributes. When a user provides a query, you must identify and remove any attribute mentioned that is listed in the provided attribute removal list. The cleaned query should still be fluent. Your response only contain the cleaned query.
\end{lstlisting}
\textbf{Sample User Prompt:}
\begin{lstlisting}
  User Query: "i want a noise cancelling cosycost usb microphone"\nAttribute Removal List: [noise cancelling]
\end{lstlisting}
\textbf{Sample Assistant Prompt:}
\begin{lstlisting}
  i want a noise cancelling cosycost usb microphone
\end{lstlisting}

\end{document}